\RequirePackage{fix-cm}
\documentclass[twocolumn,fleqn]{svjour3}          
\smartqed  
\usepackage{graphicx}
\usepackage{mathptmx}      

\usepackage{color}
\usepackage{algorithm}
\usepackage{algorithmicx}
\usepackage{amssymb}
\usepackage{CJK}
\usepackage{url}
\usepackage{ragged2e}
\usepackage{adjustbox}
\usepackage{multirow}
\usepackage[misc]{ifsym}
\usepackage{xcolor}
\usepackage{array}
\usepackage{booktabs}
\usepackage{threeparttable}
\usepackage{algpseudocode}
\usepackage{algorithm,float}
\usepackage[fleqn]{amsmath}
\definecolor{darkgreen}{RGB}{0,180,0}
\floatname{algorithm}{Algorithm}

\begin{document}
\title{VNE Strategy based on Chaotic Hybrid Flower Pollination Algorithm Considering Multi-criteria Decision Making
}
\author{Peiying Zhang\textsuperscript{1}        \and
        Fanglin Liu\textsuperscript{1}        \and
        Gagangeet Singh Aujla\textsuperscript{2,3}        \and
        Sahil Vashist\textsuperscript{4}        \and
}
\institute{\\\Letter \quad Peiying Zhang and Gagangeet Singh Aujla \\ 25640521@qq.com, gagangeet.aujla@ieee.org
    \\$^{1}$ College of Computer Science and Technology, China University of Petroleum (East China), Qingdao 266580, P.R. China
    \\$^{2}$ School of Computing, Newcastle University, Newcastle Upon Tyne, United Kingdom
    \\$^{3}$ Computer Science and Engineering Department, Chandigarh University, Gharuan, Mohali, Punjab, India
    \\$^{4}$ Computer Science and Engineering, CGC College of Engineering, Mohali, Punjab, India 140307
}
\date{Received: date / Accepted: date}
\maketitle
\begin{abstract}

With the development of science and technology and the need for Multi-Criteria Decision-Making (MCDM), the optimization problem to be solved becomes extremely complex. The theoretically accurate and optimal solutions are often difficult to obtain. Therefore, meta-heuristic algorithms based on multi-point search have received extensive attention. The flower pollination algorithm (FPA) is a new swarm intelligence meta-heuristic algorithm, which can effectively control the balance between global search and local search through a handover probability, and gradually attracts the attention of researchers. However, the algorithm still has problems that are common to optimization algorithms. For example, the global search operation guided by the optimal solution is easy to lead the algorithm into local optimum, and the vector-guided search process is not suitable for solving some problems in discrete space. Moreover, the algorithm does not consider dynamic multi-criteria decision problems well. Aiming at these problems, the design strategy of hybrid flower pollination algorithm for Virtual Network Embedding (VNE) problem is discussed. Combining the advantages of the Genetic Algorithm (GA) and FPA, the algorithm is optimized for the characteristics of discrete optimization problems. The cross operation is used to replace the cross-pollination operation to complete the global search and replace the mutation operation with self-pollination operation to enhance the ability of local search. Moreover, a life cycle mechanism is introduced as a complement to the traditional fitness-based selection strategy to avoid premature convergence. A chaotic optimization strategy is introduced to replace the random sequence-guided crossover process to strengthen the global search capability and reduce the probability of producing invalid individuals. In addition, a 2-layer BP neural network is introduced to replace the traditional objective function to strengthen the dynamic MCDM ability. Simulation results show that the proposed method has good performance in link load balancing, revenue-cost ratio, VN requests acceptance ratio, mapping average quotation, average time delay, average packet loss rate, and the average running time of the algorithm.
\end{abstract}

\section{Introduction}
\label{intro}
With the coordinated development of multiple technologies such as Internet of Things, big data, cloud computing and mobile computing, the next-generation communication network will face more challenges in the development process, such as dynamic QoS, resource vacancy, network security, and network rigidity. At present, there are many studies on these issues. For example, dynamic spectrum sensing and access technology alleviates the problem of spectrum resource shortage by utilizing spectrum holes \cite{Chunxiao6464633,Chunxiao6489503}, and the non-orthogonal multiple access technology improves resource utilization by multiplexing the power domain and the code domain \cite{Chunxiao7972935}. In addition, network virtualization, which is also a key technology of 5G, has also emerged as an effective solution to network rigidity \cite{Tutschku2009Network}. Therein, Software defined network (SDN) is a new architecture that provides a platform for network virtualization. It implements flexible control of traffic by introducing programmable controllers at the control layer. Being different from the traditional architecture, which needs to communicate and negotiate with the infrastructure providers, the SDN has the advantages of flexible deployment and quick response. In addition, the Virtual Network Embedding (VNE) is one of the key problems in network virtualization, and it has been proved to be a NP-Hard optimization problem. The research on optimization problems has been greatly promoted by the rapid development of computer technology. However, the traditional methods have obvious disadvantages when solving complex optimization problems such as objective functions and constraints that cannot be formulated. Therefore, thanks to the actual demand, many meta-heuristic intelligent algorithms based on multi-point search are proposed. However, there are still some unsolved problems in the existing heuristic algorithms \cite{a5}, such as: (1) parameter tuning is a complex and repeated process when the algorithm has many parameters. (2) the search process guided by the optimal solution is easy to lead the algorithm into a local optimum. (3) the search ability is too low when solving special optimization problems and then leads to premature convergence. (4) the existing methods of calculating fitness cannot meet the requirements of MCDM. In addition, for the VNE problem in discrete space, the optimization process guided by the direction vector is not effective. In view of these problems, a new VNE strategy considering dynamic multi-criteria decision making is proposed. The main contributions and main ideas are summarized as follows:

1. A life cycle mechanism is introduced to limit the number of iterations an individual can survive in a population. This strategy can increase the ability to acquire the lost patterns, and thus avoid premature convergence. In addition, the life span dynamically calculated based on fitness can control the death frequency of individuals in the population, so as to avoid the influence of the convergence speed due to the rapid decline of the overall quality of the population.

2. A chaos optimization strategy is introduced to replace the crossover process guided by random sequence. This strategy makes use of the excellent ergodicity of chaos sequences to enhance the global search capability of the algorithm. In addition, compared with the traditional crossover method, it can effectively avoid the generation of invalid individuals.

3. An optimized self-pollination strategy is introduced to improve the local search ability of the algorithm. Moreover, the handover probability in FPA is retained to control the balance between global and local search. In addition, a two-layer BP neural network is introduced to evaluate the performance of solutions with slight differences, which makes the local search phase have more accurate judgment.

The reminder of this paper is organized as follows. Section 2 reviews the existing methods for VNE. Section 3 introduces the network model and problem statement. Section 4 introduces the core strategies used in BP-HFPA method. In Section 5, we describe our proposed method BP-HFPA in detail. The performance of our method and other methods is evaluated in Section 6. Section 7 concludes this paper.
\section{Related Work}
\label{sec:1}
In \cite{inproceedingsth2018}, the meta-heuristic algorithms are divided into four categories: (1) based on natural evolution, (2) based on biological behavior, (3) based on physical phenomenon, and (4) based on human behavior. This section will analyze the existing meta-heuristic algorithms based on this classification strategy, and introduce the research status of VNE problem.
\subsection{Meta Heuristic Algorithms}
\label{sec:2}
(1) Natural evolution. The GA is the representative of this kind of algorithm \cite{a7,a10,h1,h2,h3,h5} In addition, the mutation based evolutionary strategy and evolutionary planning are not suitable for solving problems with abstract constraints and goals. Genetic planning is a variation of GA, the difference is that the individuals of this algorithm is a function. Differential Evolution Algorithm (DE) \cite{a8} has similar iterative steps to GA, but it relies on vector operation to guide evolution. (2) Biological behavior. The representative of this kind of algorithm is Particle Swarm Optimization algorithm (PSO) algorithm based on bird foraging behavior \cite{488968,a1,a2}. In addition, Artificial Bee Colony algorithm (ABC) can transmit the information of food source to other bees through leading bees \cite{Karaboga2007A}, Ant Colony Optimization algorithm (ACO) selects new paths through the concentration of pheromones left by other ants \cite{Dorigo2005Ant,a4}, and Firefly Algorithm (FA) guides the mutual attraction between individuals through brightness \cite{YangFirefly}. It can be seen that compared with evolution-based algorithms, individuals of this type of algorithm have certain information interaction capabilities. (3) Physical phenomenon. This kind of algorithm is relatively new. Gravity Search Algorithm (GSA) \cite{RashediGSA}, which is based on Newton's law of gravity and motion. Big Bang-Big Crunch Algorithm (BBBC) is based on the theory of cosmic explosion \cite{BBBC}. In addition, there are also Black Hole Algorithm (BHA) to simulate black hole phenomenon \cite{Kumar2015Black}, Lightning Search Algorithm (LSA) to simulate the natural phenomenon of lightning and the mechanism of step leader propagation \cite{ShareefLightning}. (4) Human behavior. This kind of algorithm simulates human behavior. For example, Harmony Search algorithm (HS) simulates the process of harmony playing \cite{Geem2009Music}. Teaching-Learning-Based Optimization (TLBO) simulates the process of class teaching \cite{RaoTeaching}. Imperialist Competitive Algorithm (ICA) simulates the annexation and competition among countries \cite{4425083}.

According to the no free lunch (NFL) theorem, no single algorithm is suitable for solving all optimization problems. As a result, universality and pertinence cannot be considered at the same time. Therefore, we believe that appropriate algorithms should be selected and optimized for specific problems.
\subsection{VNE Strategies}
\label{sec:3}
The authors of \cite{Cao2018Heuristic} divided the VNE methods into two types: optimal algorithms and heuristic algorithms. In recent studies on optimal algorithm, the authors of \cite{articleC} proposed a dynamic VNE algorithm that can meet the requirements of customized QOS. In the same year, the authors of \cite{DBLP:conf/icc/CaoGWQZY19} proposed a VNE algorithm based on five constraints, and the authors of \cite{articleFF} further considered the perception of energy consumption. Moreover, the authors of \cite{articleHH} proposed a candidate set based VNE algorithm considering the time delay and geographical location constraints. The authors of \cite{articleAAA} proposed a method comprehensively considered five topology attributes and global network resources. However, in the case of large problem scale, solving the optimal solution often consumes a lot of computing resources. As a result, more researches has focused on the heuristic algorithms with lower computation. In the early research \cite{Cheng2011Virtual}, PSO algorithm was used to solve the VNE problem for the first time. In follow-up studies, the authors of \cite{Song2019Distributed} proposed a distributed VNE algorithm (HA-VNE-PSO) which can skip unnecessary iterative steps. The authors of \cite{Zhang2013A} proposed a unified enhanced VNE algorithm (VNE-UEPSO) based on PSO. The authors of \cite{8695261} proposed a VNE algorithm based on a candidate point selection strategy to make up for the shortcomings of heuristic solution was not accurate enough. In addition, there are also many researches based on GA. For example, the authors of \cite{Zhuang2018A} proposed a VNE algorithm based on cellular genetic mechanism, which further improved the interaction ability of the population. Moreover, the authors of \cite{6201795} combined the annealing algorithm with GA to improve the acceptance rate and reduce the cost, and the authors of \cite{7943336} discussed the problem of using GA to deal with the multi-virtual network embedding (MVNE).

It is worth noting that the objective function of the above meta-heuristic algorithm only considers resources or prices, and can only preliminarily distinguish the quality of different solutions. Therefore, further optimization is needed.
\section{Network Model and Problem Statement}
\begin{figure}[htbp]
\small
\centering
\includegraphics[width=8cm]{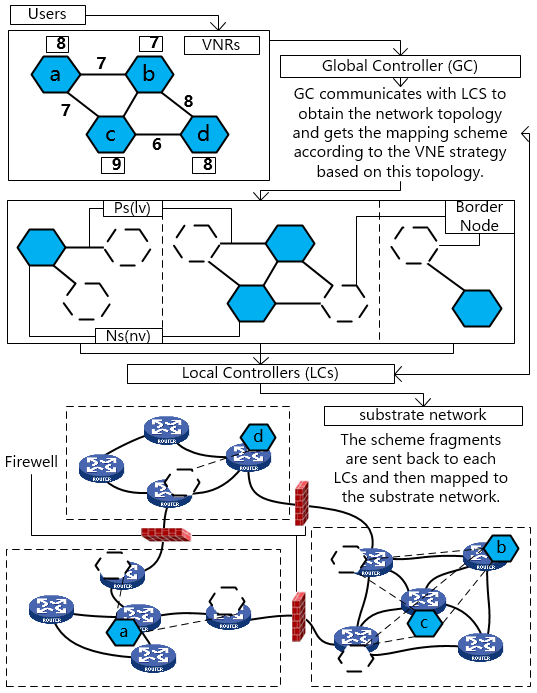}\\
\caption{Process of solving the VNE problem.}
\label{fig:1}
\end{figure}

\begin{figure}[htbp]
\small
\centering
\includegraphics[width=8cm]{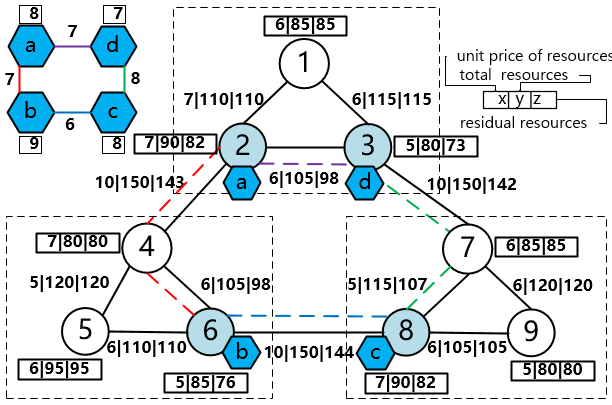}\\
\caption{An substrate network consisting of three domains and a VNR.}
\label{fig:2}
\end{figure}

Figure.~\ref{fig:1} is the process of solving the VNE problem, and it can be simplified as Fig.~\ref{fig:2}. The top left of Fig.~\ref{fig:2} is a VN to be mapped, and the right is a substrate network topology composed of three domains. In this paper, the Fig.~\ref{fig:2} will be used as references.
\subsection{Substrate Network and Virtual Network Model}
In this model, the substrate network is defined as an undirected graph $G^s=\{N^s,L^s\}$, where the $N^s$ represents a set of substrate nodes and the $L^s$ represents a set of substrate links. $\{CPU(n_s),UP(n_s),TD(n_s),PLR(n_s)\}$ represents a set of attributes of each substrate node, which respectively represent the CPU capacity, the resource unit price, the time delay within substrate nodes, and the packet loss rate. Moreover, $\{BW(l_s),UP (l_s),AUP (P_s),TD (l_s)\}$ represents a set of attributes of each substrate link, which respectively represent the bandwidth, the resource unit price, the aggregation unit price, and the time delay of substrate links. Therein, the $AUP (P_s)$ can be expressed as follows:
\begin{equation}
AUP(P_s)=\sum_{l_s\in P_s}UP(l_s),
\end{equation}
where the $P_s$ represents a substrate path composed of several substrate links. A VN can also be abstracted as a undirected graph $G^v = \{N^v,L^v\}$, the $N^v$ represents a set of virtual nodes and the $L^v$ represents a set of virtual links. The $CPU(n_v)$ represents the CPU capacity required by the virtual node $n_v$, and the $BW(l_v)$ represents the bandwidth required by the virtual link $l_v$.
\subsection{Virtual Network Embedding Problem Description}
The process can be modeled as $M:G^v\{N^v, L^v\}\to G^s\{N^s, L^s\}$. Each virtual node $n_v\in N^v$ chooses a substrate node that conforms to the constraint condition as the mapping target. It should be noted that different virtual nodes in the same VNR cannot be mapped repeatedly to the same substrate node. Each virtual link is mapped to a substrate path $P_s$ that conforms to the constraint condition, where the endpoints of the path are determined during the node mapping phase. In addition, the constraints can be defined as:
\begin{equation}
\begin{cases}
BW(l_v)\leq BW(l_s),\;\forall l_s\in Ps(l_v),\\
CPU(n_v)\leq CPU(n_s),\;n_v\leftrightarrow n_s,
\end{cases}
\end{equation}
where the $\leftrightarrow$ represents the process of mapping $n_v$ to $n_s$. In addition, a scheme for mapping VN can be expressed as $M(G^v,N^s,P^s)$, where the $N^s$ and the $P^s$ respectively represent the set of substrate nodes and the set of substrate paths in this scheme.
\subsection{Objectives and Evaluation Index}
The objective function can be expressed as:
\begin{equation}
\begin{aligned}
OBJ(G^v) = &min\sum_{n_v\in N^v}CPU(n_v)\times UP(n_s)+\\
&\sum_{l_v\in L^v}BW(l_v)\times AUP(P_s).
\end{aligned}
\end{equation}

The link load balancing can be measured by the variance of consumed resources, and it can be expressed as:
\begin{equation}
\sigma^2=\frac{\sum_{l_s\in L^s}(used\;resources\;for\;l_s-\mu)^2}{num(L^s)},
\end{equation}
where the $\mu$ represents the average used resources, and the $num(L^s)$ is the number of links in the substrate network.

The revenue-cost ratio can be expressed as $(revenue(G^v)$ $/cost(G^v))$, where the revenue and the cost of mapping a VN can be expressed as:
\begin{equation}
revenue(G^v)=\sum_{n_v\in N^v}CPU(n_v)+\sum_{l_v\in L^v}BW(l_v),
\end{equation}
\begin{equation}
cost(G^v)=\sum_{n_v\in N^v}CPU(n_v)+\sum_{l_v\in L^v}BW(l_v)Hops(P_s(l_v)),
\end{equation}
where the $Hops(P_s(l_v))$ represents the number of hops of the substrate path $P_s(l_v)$.

The VN request acceptance ratio can be expressed as:
\begin{equation}
acceptance\;ratio=\frac{num(VNR_{accept})}{num(VNR_{refuse})},
\end{equation}
where the $num(VNR_{accept})$ represents the number of VNRs that were accepted and successfully mapped to the substrate network, and the $num(VNR_{refuse})$ represents the number of rejections.

The average quotation can be expressed as:
\begin{equation}
average\;quotation=\frac{\sum OBJ(G^v)}{num(VNR_{accept})}.
\end{equation}

The total time delay includes the total delay of links and the total delay of nodes, and it can be expressed as:
\begin{equation}
TD(M(G^v,N^s,P^s))=\sum_{n_s\in N^s}TD(n_s)+\sum_{P_s\in P^s}\sum_{l_s\in P_s}TD(l_s).
\end{equation}

The average time delay can be expressed as:
\begin{equation}
average\;delay=\frac{\sum TD(M(G^v,N^s,P^s))}{num(VNR_{accept})}.
\end{equation}

The total packet loss rate can be expressed as:
\begin{equation}
PLR(M(G^v,N^s,P^s))=\sum_{n_s\in N^s}PLR(n_s).
\end{equation}

The average packet loss rate can be expressed as:
\begin{equation}
average\;PLR=\frac{\sum PLR(M(G^v,N^s,P^s))}{num(VNR_{accept})}.
\end{equation}

The average running time is the average time consumed by mapping a VN, and in milliseconds.
\section{Strategy Model and Innovation Motivations}
\subsection{Life Cycle Mechanism}
\begin{figure}[htbp]
\small
\centering
\includegraphics[width=8cm]{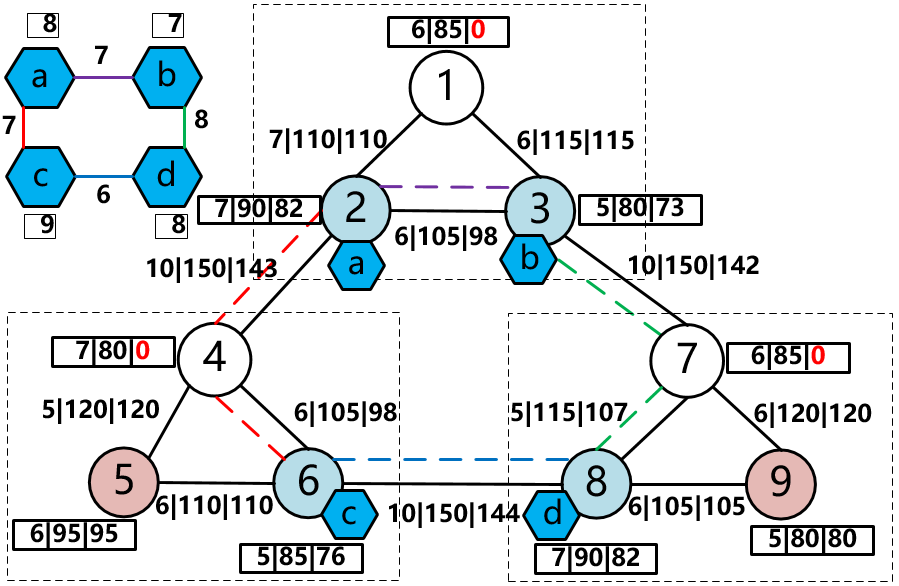}\\
\caption{An substrate network consisting of three domains and a VNR.}
\label{fig:15}
\end{figure}

The authors of \cite{articlePPP} explained that the cause of premature convergence is pattern loss. Moreover, due to the search characteristics of GA, lost patterns can only be retrieved during the mutation phase. However, the mutation of small probability is not enough to deal with large-scale pattern loss. In addition, because of the characteristics of VNE problem, the lost patterns will be more difficult to obtain, and as shown in the Fig.~\ref{fig:15}. The nodes marked in red in Fig.~\ref{fig:15} are nodes that can be selected as mutation targets. It can be seen that because different virtual nodes in the same VN cannot be mapped to the same substrate node, there are very few optional nodes. This is especially true when there are fewer resources on the substrate network. Therefore, even if a lost pattern is acquired through mutation, another pattern may be lost because there are too few optional nodes.

In order to solve this problem, we introduce a life cycle mechanism to limit the number of iterations for individuals to survive in the population, to avoid obsolete individuals affecting the genetic diversity of the population. Compared to other similar strategies that are aperiodic and random, this strategy can gradually replace the individuals that have existed for a long time. Therefore, it can take into account the overall quality of the population while avoiding premature convergence, thereby ensuring that the convergence rate is not affected. We define the life cycle as the following expression:
\begin{equation}
life\;cycle=
\begin{cases}
\lambda_{life}F(x_i)\frac{num(X)}{\sum_{x_i\in X}F(x_i)}\times5,M < 25,\\
\lambda_{life}F(x_i)\frac{num(X)}{\sum_{x_i\in X}F(x_i)}\frac{M}{5},M\geq25,
\end{cases}
\end{equation}
where the $\lambda_{life}$ represents the life factor, which ranges from 0 to 3, and the default is 1. The $F(x_i)$ represents the fitness of the solution $x_i$, the $num(X)$ represents the number of individuals in the population $X$, and the M represents the maximum number of iterations. In addition, a method was introduced to further enhance the ability to capture lost patterns. When a new individual is used to replace the old individual, a 0-1 sequence is obtained by bitwise-and operation of new and old individuals. Finally, when a component in the 0-1 sequence is 1, the rest of the substrate nodes that satisfy the constraint will be randomly selected as the corresponding component in the new individual.
\subsection{Chaos  Strategy}
After several iterations, it is clear that some nodes that can reduce the fitness of the individuals will repeatedly appear in the descendants that are retained. In which case, the traditional crossover strategy is likely to produce invalid individuals, as shown in the Fig.~\ref{fig:14}. Take single-point crossover as an example. If the substrate node N exists simultaneously in a pair of selected individuals A and B to be crossed, invalid individuals will be generated after crossover when the node N is in the same color region. By calculation, the probability of N in the same color region is 50\%. Therefore, the probability of producing invalid individuals after crossover is also 50\%. This will obviously reduce the search ability of crossover process greatly.
\begin{figure}[htbp]
\small
\centering
\includegraphics[width=8cm]{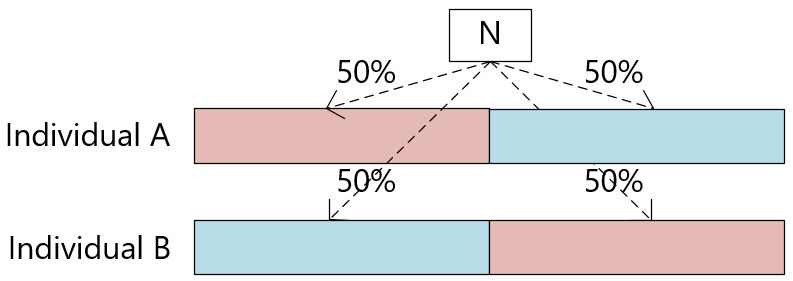}\\
\caption{A diagram of single point crossing.}
\label{fig:14}
\end{figure}

To solve this problem, we introduce a chaos optimization strategy \cite{a9} into GA and use its excellent ergodicity to guide the crossover process. In this paper, we define a chaos sequence generator based on Logistic model and convert the obtained chaos sequence into a mask code for crossover, as follows:
\begin{equation}
 x_{n+1}=ux_n(1-x_n),u\in[0,4],x\in(0,1),
\end{equation}
when $u$ is greater than 4, the model will diverge, that is, $x$ is greater than 1. But when it is less than or equal to 4, with the increase of $u$, the model will appear period doubling bifurcation until it shows a random characteristic, namely chaos. Because we want to get a sequence which is almost random but has chaotic characteristics, we set $u$ to 4 and get the initial $x$ randomly. Define the mask code for crossover as $M[m_1,m_2,..., m_k]$, where $k$ is equal to the dimension of the solution, and use the following expression to convert the chaos sequence to the mask code:
\begin{equation}
m_i=
\begin{cases}
0, x_i < 0.5,\\
1,x_i\geq0.5,
\end{cases}
\end{equation}
therein, when the value is 1, the parents' genes in this dimension will be crossed, and when the value is 0, the genes will not be crossed.
\subsection{Self Pollination Strategy}
\begin{figure}[htbp]
\small
\centering
\includegraphics[width=8cm]{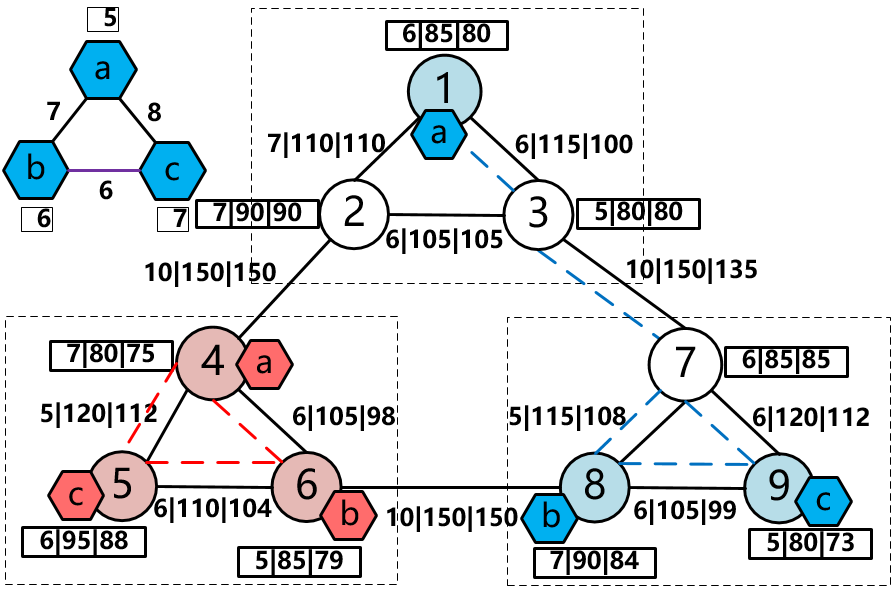}\\
\caption{A self-pollination process.}
\label{fig:13}
\end{figure}

It is well known that traditional mutation process are not enough to deal with the large-scale pattern loss. Moreover, the combination of life cycle strategy and chaos optimization strategy makes the algorithm difficult to fall into local optimal solution. Therefore, the optimized algorithm does not need the traditional mutation operation any more. At the same time, considering that the algorithm with enhanced global search ability has significantly weaker local search ability, we introduce the self-pollination operation from the FPA to replace the traditional mutation operation. Therein, when the number of iterations is t, the self pollination process can be defined as follows:
\begin{equation}
 x^{t+1}_i=x^t_i+\varepsilon(x^t_j-x^t_k),\varepsilon\in[0,1],
\end{equation}
where $x^t_i$ represents the individuals to be self pollinated, $x^t_j$ and $x^t_k$ represent a pair of randomly selected individuals, and $\varepsilon$ is a random number uniformly distributed in the range of 0 to 1. In view of VNE problem, a common problem is that individuals cannot control the local search within a certain range by adding with an unprocessed vector. For example, define two individuals $x^t_j$[1,3,6] and $x^t_k$[4,1,2], and a individual $x^t_i$[4,6,5] to be self-pollinated. Moreover, considering the discrete characteristics of VNE problem, the $\varepsilon$ is set to 1. According to $\varepsilon(x^t_j-x^t_k)$ can get the pollen $\nu$[-3,2,4], then the individual after self-pollination is $x^{t+1}_i$[1,8,9]. This process is shown in Fig.~\ref{fig:13}.
It can be seen in the Fig.~\ref{fig:13} that the individual $x^{t+1}_i$ is far away from the individual $x^t_i$ in geographical location. Therefore, this process does not reflect the characteristics of local search. To solve this problem, we define $\varepsilon$ as a process, which is expressed as follows:
\begin{equation}
d^{new}_i=
\begin{cases}
1, d_i > 0,\\
0, d_i = 0,\\
-1, d_i < 0,
\end{cases}
\end{equation}
where $d_i$ is the component of pollen vector $\nu$. Because the number of substrate nodes is continuous in the same domain, this method can allow individuals search locally in a small range at the geographic location level, as shown in Fig.~\ref{fig:14}.
\begin{figure}[htbp]
\small
\centering
\includegraphics[width=8cm]{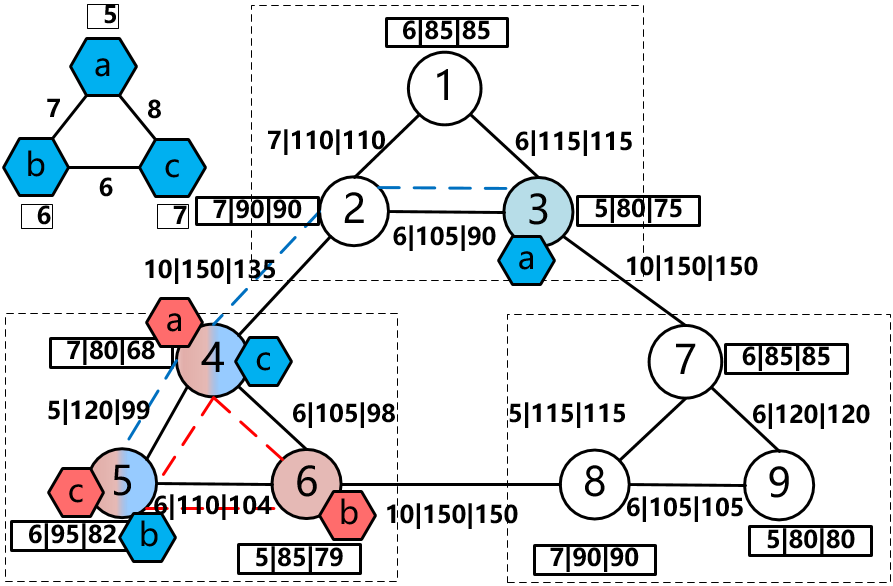}\\
\caption{A self-pollination process using new strategy.}
\label{fig:14}
\end{figure}
After being treated by $\varepsilon$, pollen $\nu$[-3,2,4] becomes $\nu$[-1,1,1], while individual $x^t_i$[4,6,5] becomes $x^{t+1}_i$[3,5,4] after pollination. It can be seen that the new strategy can reflect the characteristics of local search. In addition, we use the transfer probability $p_t$ to control the switching between global search and local search, $p\in[0,1]$.
\subsection{BP Neural Network}
The evolutionary direction of traditional meta-heuristic algorithms is usually guided by an expression called fitness function ($F(x_i)$), which is usually the objective function multiplied by a conversion coefficient, as follows:
\begin{equation}
F(x_i)=conversion\;coefficient \times OBJ(G^v).
\end{equation}

\begin{figure*}[htbp]
\small
\centering
\includegraphics[width=15cm]{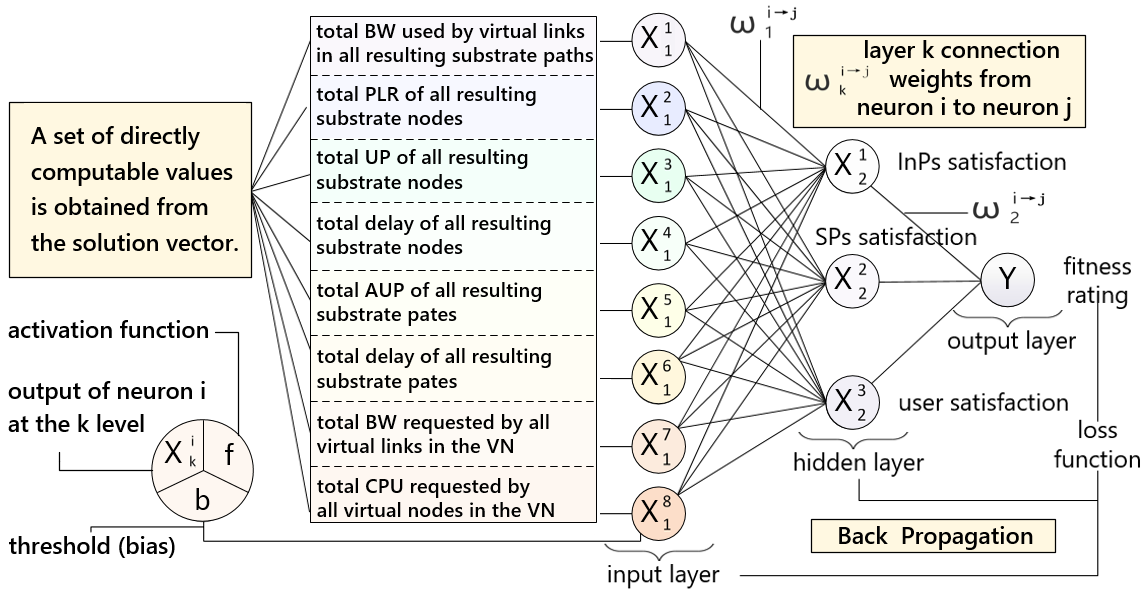}\\
\caption{A two-layer BP neural network for fitness rating.}
\label{fig:16}
\end{figure*}

However, as the scale of the problem increases, the conventional fitness function based on experience is obviously insufficient to satisfy the consideration of MCDM. In contrast, it is easier to give a fitness rating (poor, average, or good) based on actual use results or experience than to give a specific fitness value. If a set of fitness ratings is regarded as a set of discrete solutions of an undetermined fitness function, it is a problem to be considered to choose a appropriate fitting method to obtain the coefficients of this function. Because machine learning is a better choice for solving complex fitting problems than manual adjustment \cite{Jiang7792374}, a two-layer BP neural network is introduced to assist the calculation of fitness \cite{RUMELHART1986Learning,a3,a6}, as shown in the Fig.~\ref{fig:16}.
Therein, the input value of each neuron in the input layer and the corresponding calculation method are highlighted in the same color. Moreover,the ReLU proposed in paper \cite{Krizhevsky2012ImageNet} is used as the activation function. In addition, the loss function is used to abstract the difference between the output value and the standard value of each neuron into the loss factor, and it can be defined as:
\begin{equation}
\delta_{output}=y_{out}\times(1-y_{out})\times(y_{out} - y_{true}),
\end{equation}
\begin{equation}
\delta_{hidden}^k=x_k^i\times(1-x_k^i)\times\sum_{n=1}^m\omega_k^{j\to i}\delta_{k+1}^n,
\end{equation}
where $\delta_{output}$ represents the error factor of neurons feedback in output layer, $\delta_{hidden}^k$ represents the error factor of neurons feedback in hidden layer k, $y_{out}$ represents the output value, $y_{true}$ represents the true value, $\delta_{k+1}^n$ represents the error factor of neurons $n$ feedback in the next layer, m is equal to the number of neurons in the next layer, and other information that may be needed is marked on the Fig.~\ref{fig:16}. In addition, $y$ represents the output value of the neuron in the output layer, which is calculated by the output value of the neuron in the previous layer, and can be expressed for the following equation:
\begin{equation}
\begin{split}
y=&f_{BPnn}(x_i)\\
=&ReLU(\vec{\omega_2}\cdot\vec{x_2^T}+b),
\end{split}
\end{equation}
where $\vec{\omega_i}$ represents the row vector of the weight of layer i, $\vec{x_i}$ represents the column vector of the neuron output value of layer i, and $b$ represents the threshold (or bias), which measures how easy it is for a neuron to generate excitation.

BP neural network can input a set of data samples with known ratings, calculate the loss $\delta$ for all neurons, and then adjust the weight and threshold value of the network through the back propagation process, which is called supervised learning. The adjust method can be formulated as:
\begin{equation}
\omega_k^{i\to j}=\omega_k^{i\to j}+\eta\times\delta_k^j\times\omega_k^{i\to j},
\end{equation}
\begin{equation}
b_k^i=b_k^i+\eta\times\delta_k^i,
\end{equation}
where $\eta$ represents a learning factor, a predefined constant used to adjust the learning speed, $\delta_k^j$ represents the loss factor of the reverse output of the neuron $j$ at the $k$ layer, and $b_k^i$ represents the bias of the neuron $i$ at the $k$ layer. Moreover, the initial values of the weights and thresholds follow a normal distribution where the mean is 0 and the variance is 1. In addition, since some steps in the BP-HFPA algorithm depend on specific fitness values, so the fitness level should be appropriately processed, and it can be formulated as:
\begin{equation}
fitness=\lambda_{user}\frac{f_{BPnn}(x_i)}{n}\times F(x_i) + (1-\lambda_{user})F(x_i),
\end{equation}
where $\lambda_{user}$ represents the importance of the user and ranges from 0 to 1, $n$ is equal to the number of ratings, and the ratings can be expressed as integers 1 to $n$, and the smaller the better.
\section{Heuristic Algorithm Design}
\subsection{Node Mapping Algorithm}
We use a hybrid algorithm of GA and FPA to complete node mapping. The evolution process of traditional genetic algorithms usually includes selection, crossover and mutation stages. In this model, the elite selection strategy is adopted to retain half of the individuals with lower fitness, a chaotic sequence is used to guide the crossover process, and the mutation phase is replaced by the self-pollination process in the FPA. In addition, the individuals are encoded as $X_i=\{X_i^{1\to j_1},X_i^{2\to j_2},...X_i^{k\to j_k}...X_i^{n\to j_n}\}$. The $X_i$ represents the mapping strategy of VN with n virtual nodes, and the $x_i^{k\to j_k}$ represents the substrate node to which the virtual node $k$ in the VN is mapped. Therein, $j_k$ is equal to the number of this substrate node, and that the substrate node $j_k$ should satisfy the constraints of the virtual node $k$. Moreover, different fitness calculation methods will be used at different phases. In the global search phase, the fitness function $F(x_i)$ is used to calculate the fitness of individuals. In the local search (self-pollination) phase, the trained two-layer BP neural network is used to calculate the fitness, where the calculation method has been given in Section 4.4. In order to avoid the offsprings generated by crossover are not feasible solutions of the VNE problem, a feasibility judgment is added.

The detailed steps of node mapping algorithm are illustrated in Algorithm 1.
    \begin{algorithm}
        \caption{The Node Mapping Algorithm Based on Hybrid Flower Pollination Algorithm.}
        \begin{algorithmic}[1] 
            \Require $G^s = \{N^s,L^s\}$ and $G^v = \{N^v,L^v\}$.
            \Ensure $M(G^v,N^s,P^s)$, where $P^s$ is not determined.
            \State $P_t$ $\gets$ transfer probability;
            \State $P_c$ $\gets$ cross probability;
            \State X $\gets$ maximum population capacity;
            \State M $\gets$ maximum iterations;
            \State Randomly generate X individuals;
            \For{not reached M iterations}
                \State Life judgment and add new individuals;
                \If {random decimal $ > P_t$}
                    \State Select $\frac{X}{2}$ individuals;
                    \While{the number of individuals is less than X}
                        \State Select a pair of individual at random;
                        \If {random decimal $ < P_c$}
                            \State Crossing this two individuals;
                        \EndIf
                        \State Feasibility judgment;
                    \EndWhile
                    \State Reduced parental lifespan;
                \Else
                    \State Self-pollination for the whole population;
                    \State Reduces the lifespan of unrenewed individuals;
                    \State Resets the lifespan of the updated individuals;
                \EndIf
            \EndFor
            \State \Return{The individual with the lowest fitness;}
        \end{algorithmic}
    \end{algorithm}
\subsection{Link Mapping Algorithm}
The step of weight update in Algorithm 2 refers to updating the weight of the substrate links based on a certain policy. Therein, the weight of the substrate links whose remaining resources are less than the virtual links requirement will be modified to zero. In addition, the weight of the remaining substrate links will be adjusted according to a load balancing strategy. The model uses link resource unit price as link weight, and the load balancing strategy can be expressed for the following equation:
\begin{equation}
\omega(l_s)=\left\{
\begin{array}{rcl}
UP(l_s)(1+\lambda_{weight}\times \omega_{extra}) & & {Used(l_s) > \overline{Used},}\\
UP(l_s)& & {Used(l_s) \leq \overline{Used}.}\\
\end{array} \right.
\end{equation}
\begin{equation}
\omega_{extra}=\frac{Used(l_s)-\overline{Used}}{max\{\forall l_s\in L_s\vert Used(l_s)\}-\overline{Used}},
\end{equation}
\begin{equation}
\overline{Used}=\frac{\sum_{l_s\in L^s}Used(l_s)}{num(l_s)},
\end{equation}
\begin{equation}
Used(l_s)=\sum_{l_v\in M(l_s)}BW(l_v),
\end{equation}
where the range of $\lambda_{weight}$ is (0,2], the $num(l_s)$ represents the number of substrate links, and the $M(l_s)$ represents a collection of mapped virtual links on a substrate link $l_s$.
\begin{algorithm}
        \caption{The Link Mapping Algorithm Based on Shortest Path Algorithm.}
        \begin{algorithmic}[1] 
            \Require Virtual network node mapping scheme.
            \Ensure Virtual network link mapping scheme.
            \State Sort the virtual links by the required bandwidth in nonincreasing order;
            \For{all the unmapped virtual links in VNR}
                \If {the existing scheme of this $l_v$ can be mapped}
                    \State Store the existing scheme;
                \Else
                    \State Gets the corresponding two substrate endpoints;
                    \State Update the weight of each substrate link;
                    \State Obtain the shortest path between the two endpoints;
                    \State Restoring weight;
                \EndIf
            \EndFor
            \State \Return{Link mapping scheme;}
        \end{algorithmic}
    \end{algorithm}
\section{Performance Evaluation}
This section compared the performance of four VNE methods, including VNE-CGA, MD-PSO, CAN-A, and BP-HFPA. The first two and our methods are meta-heuristics, and the third is the optimal algorithm. Therein, the VNE-CGA algorithm used the GA algorithm optimized by cellular automata to obtain the node mapping scheme. The MD-PSO algorithm used the PSO algorithm to obtain the node mapping scheme, and a candidate point selection strategy was introduced to initially screen the substrate nodes. Both algorithms used the shortest path algorithm to obtain the link mapping scheme. In addition, the CAN-A algorithm will obtain the optimal solution based on the objective function in the substrate node set and the substrate link set after preliminary screening.
\subsection{Environment Settings and Algorithm Parameters}
The substrate network topology and virtual network request topology are generated by the GT-ITM \cite{Zegura1996How} tool. The number of domains is 4, the number of nodes in a domain is 30, the CPU range of substrate nodes is 100 to 300, the TD of links and nodes is 1 to 5, the PLR range of nodes is 0.01 to 0.5, the UP range of BW and CPU is 1 to 10, the BW range of inter-domain links is 1000 to 3000, the BW range of cross-domain links is 3000 to 6000, the connection probability between nodes is 0.5, the number range of virtual nodes in a VN is 5 to 10, the requested CPU and BW range is 1 to 10, the number of VNRs in 100 time units obeys the poisson distribution with a mean value of 10, and the lifetime of a VN is 1000. In addition, about algorithm parameters, the life weight $\lambda_{life}$ is set to 1, the transfer probability $p$ is set to 0.7, the conversion coefficient of fitness function is set to 1, the user factor $\lambda_{user}$ is set to 0.7, and the intervention weight $\lambda_{weight}$ of link load balancing strategy is set to 2.
\subsection{Evaluation Results}
\begin{figure}[htbp]
 \center{\includegraphics[width=8.2cm]  {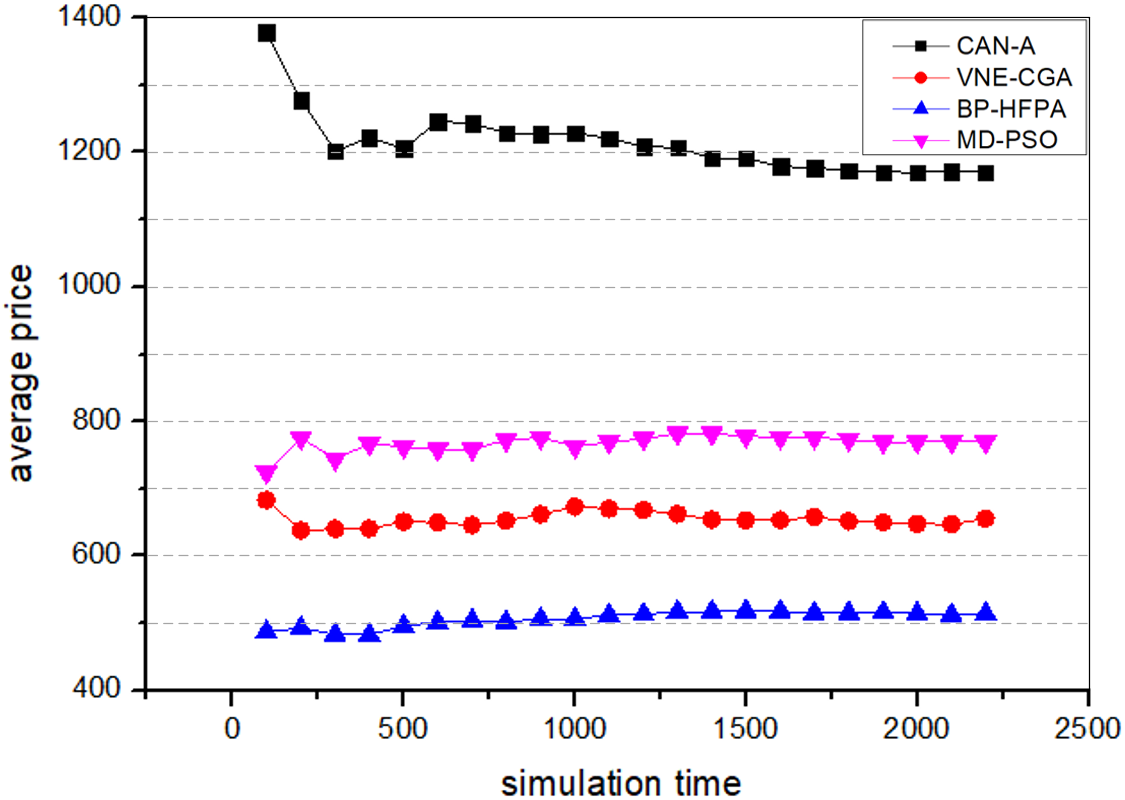}}
 \caption{The diagram of the average quotation.}
  \label{fig:6}
\end{figure}
\begin{figure}[htbp]
 \center{\includegraphics[width=8.2cm]  {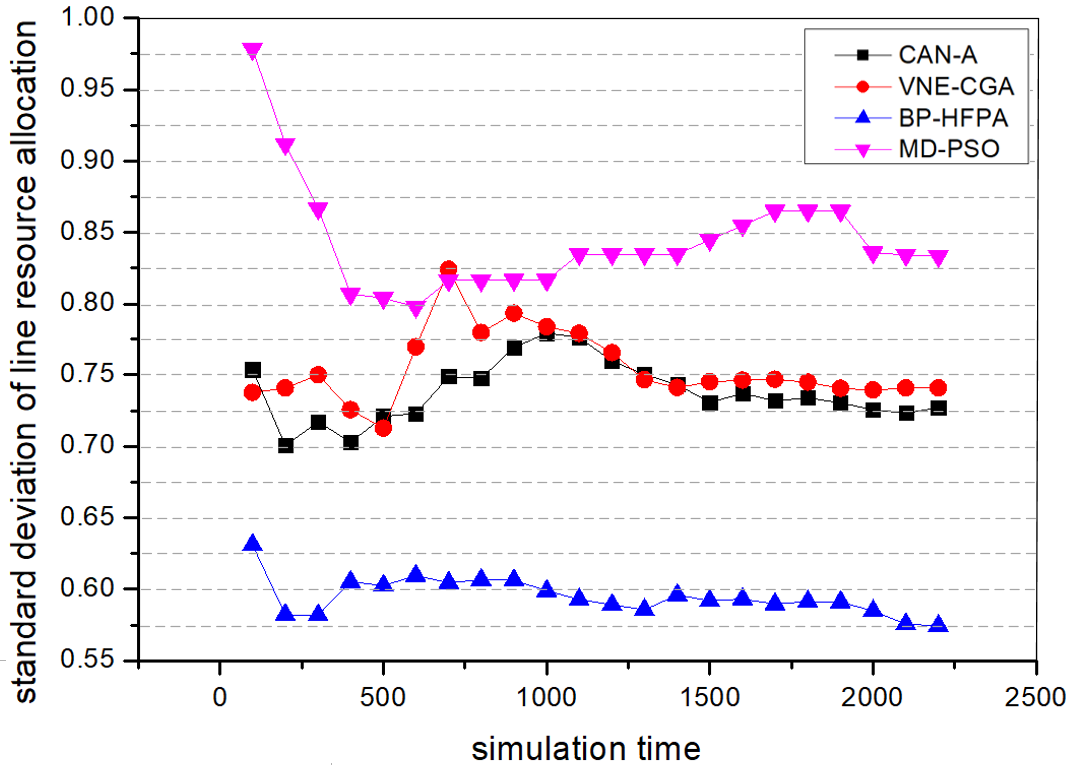}}
 \caption{The diagram of load balancing of the substrate links.}
  \label{fig:3}
\end{figure}
As can be seen from the Fig.~\ref{fig:6}, the BP-HFPA algorithm has the best performance in average mapping quotation. This is because the BP-HFPA algorithm has excellent global search ability in the crossover process guided by chaotic sequence, while the algorithm has precise local search ability in the self-pollination process guided by neural network. Therefore, the algorithm can approximate the optimal solution within a limited number of iterations and finally obtain excellent and stable results.

As can be seen from the Fig.~\ref{fig:3}, the BP-HFPA algorithm performs best in link load balancing. This is because, although the four algorithms using the shortest path algorithm for mapping virtual links, BP-HFPA algorithm considers the link load balancing. In addition, although the CAN-A algorithm takes link load balancing into consideration when generating the candidate path set, however, because time delay is also taken into account, the load balancing situation after comprehensive consideration is not as good as our algorithm.
\begin{figure}[htbp]
 \center{\includegraphics[width=8.2cm]  {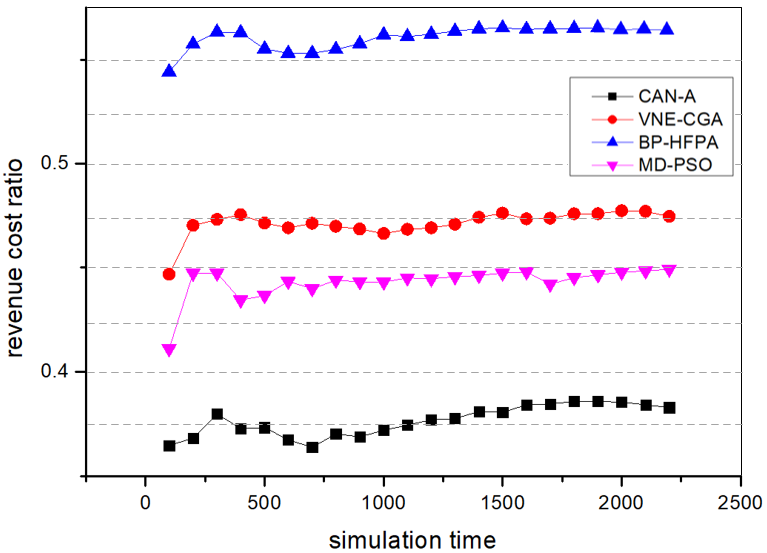}}
 \caption{The diagram of revenue-cost ratio.}
  \label{fig:4}
\end{figure}

As can be seen from the Fig.~\ref{fig:4}, the BP-HFPA algorithm performs best in terms of the revenue-cost ratio. This is because the BP-HFPA algorithm is able to obtain solutions that take into account a variety of indicators, so it also takes into account resource consumption. The CAN-A algorithm focus on delay and load balancing, the MD-PSO focus on quotation, so they perform relatively poorly.
\begin{figure}[htbp]
 \center{\includegraphics[width=8.2cm]  {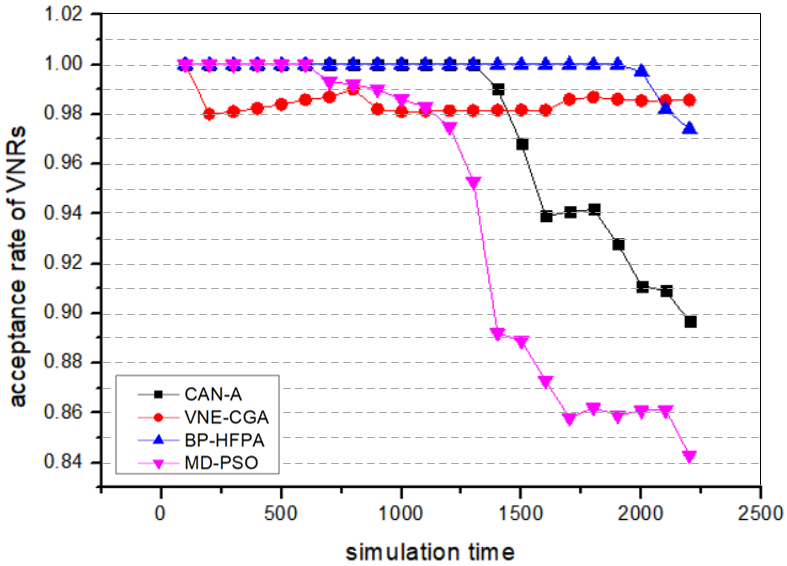}}
 \caption{The diagram of the VNRs acceptance ratio.}
  \label{fig:5}
\end{figure}

It can be seen from the Fig.~\ref{fig:5} that the BP-HFPA algorithm performs best in the VNRs acceptance rate. This is because the BP-HFPA algorithm adds a weight update operation and a remapping operation, thus avoiding most mapping failures.
\begin{figure}[htbp]
 \center{\includegraphics[width=8.2cm]  {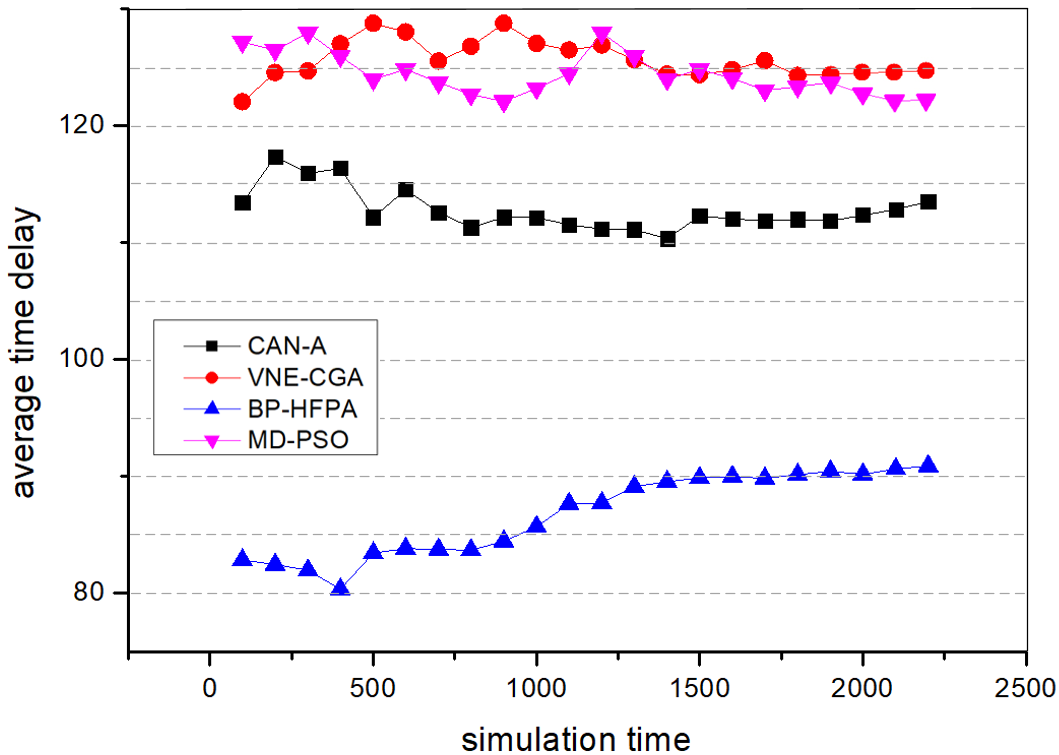}}
 \caption{The diagram of the average time delay.}
  \label{fig:7}
\end{figure}

\begin{figure}[htbp]
 \center{\includegraphics[width=8.2cm]  {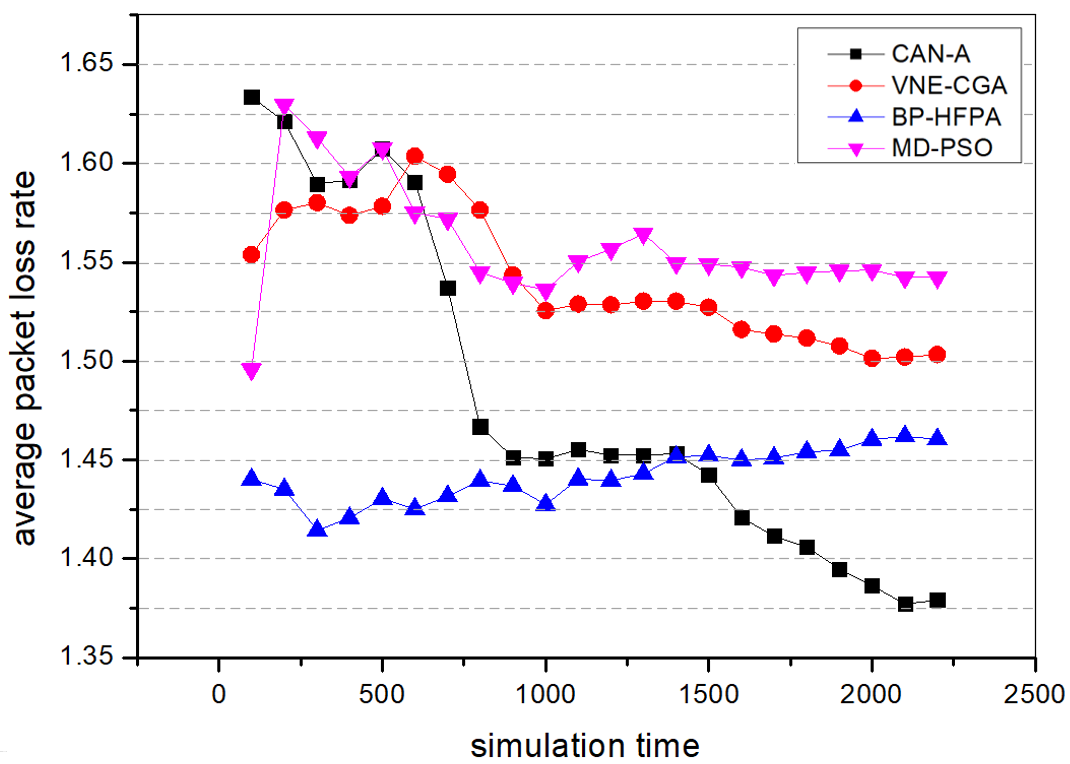}}
 \caption{The diagram of the average packet loss rate.}
  \label{fig:8}
\end{figure}

As can be seen from the Fig.~\ref{fig:7} and Fig.~\ref{fig:8}, the BP-HFPA algorithm has the best performance in terms of average packet loss rate and average delay. This is because the BP-HFPA algorithm can comprehensively score the scheme, and its excellent search ability enables it to approach the current optimal solution every time. In addition, although delay optimization \cite{h4,h6} is considered in the objective function of the CAN-A algorithm, its ability to reduce delay is still lower than that of BP-HFPA algorithm. This is because CAN-A removes a large number of solutions through screening, then better solutions may be missed. In addition, the BP-HFPA algorithm has excellent search capabilities, so it can approximate the optimal solution in a wide solution space as much as possible. Therefore, the BP-HFPA algorithm has more advantages.

\begin{figure}[htbp]
 \center{\includegraphics[width=7.8cm]  {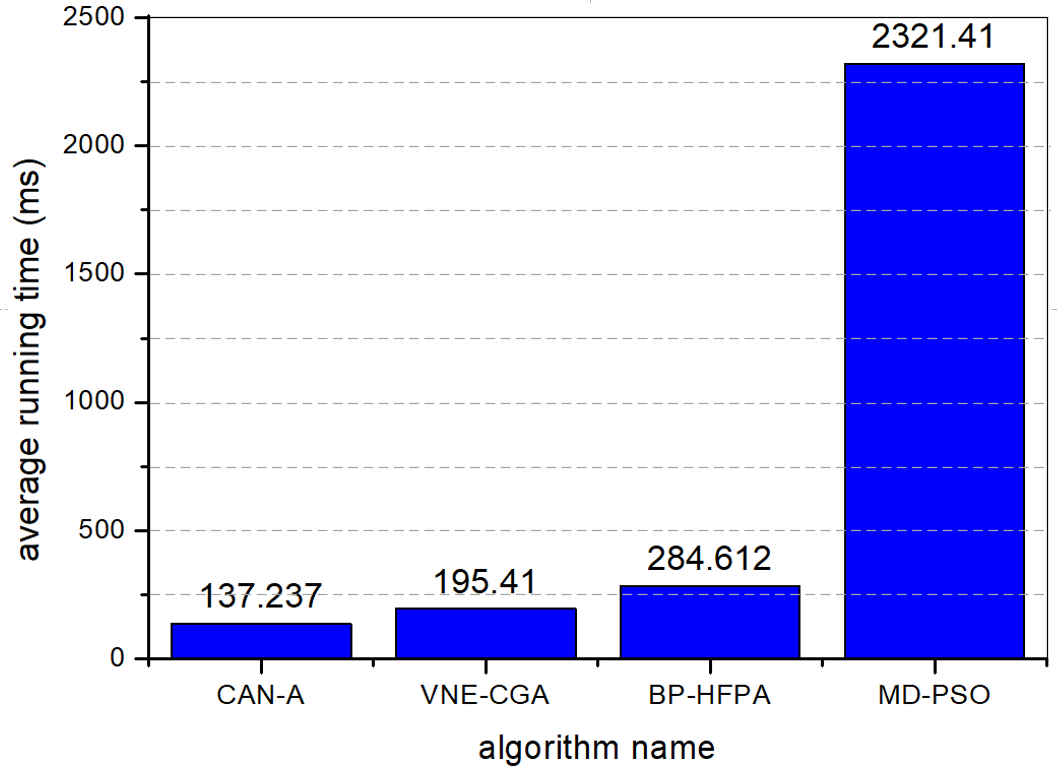}}
 \caption{The diagram of the average running time.}
  \label{fig:10}
\end{figure}

It can be seen from the Fig.~\ref{fig:10} that the average running time of the BP-HFPA algorithm is not much different from the VNE-CGA and CAN-A algorithm, which shows that although our algorithm increases a variety of search strategies, the running time is still within the acceptable range.
\section{Conclusion}
Since VNE is a NP-hard problem, a large number of researches focus on meta-heuristic algorithms. However, the existing meta-heuristic algorithms still need to be optimized when solving a VNE problem with discrete characteristics. For example, (1) the global search guided by the global optimal solution is easy to fall into the local optimal solution. (2) the search process guided by the vector cannot reflect the directionality of the vector, and it is easy to produce a large number of invalid individuals. (3) the conventional local search method by controlling step length is not suitable for VNE problem. In addition, MCDM is also worth considering. To solve these problems, this paper discusses the design of hybrid meta heuristic algorithm. Combining the advantages of GA algorithm and FPA algorithm, and a chaos optimization strategy is introduced to further enhance the search ability of the algorithm. In addition, a life cycle mechanism is introduced to avoid premature convergence. In addition, a method of fitness calculation based on two-layer BP neural network is proposed, which takes into account multiple targets and makes the comparison of similar individuals more accurate during the local search process. Simulation results show that this method has good performance in many aspects.

In future work, we will make efforts to solve the VNE problem in some complex cases, and comprehensively consider the divisibility of nodes and links and information security issues.
\begin{acknowledgements}
This work is supported by "the Fundamental Research Funds for the Central Universities" of China University of Petroleum (East China) (Grant No. 18CX02139A), the Shandong Provincial Natural Science Foundation, China (Grant No. ZR2014FQ018), and the Demonstration and Verification Platform of Network Resource Management and Control Technology (Grant No. 05N19070040). The authors also gratefully acknowledge the helpful comments and suggestions of the reviewers, which have improved the presentation.
\end{acknowledgements}

 \bibliographystyle{spmpsci}
\bibliography{references}
\end{document}